\pdfoutput=1

\documentclass[11pt,table,xcdraw]{article}

\usepackage{ACL2023}
\usepackage{times}
\usepackage{latexsym}

\usepackage[T1]{fontenc}
\usepackage[utf8]{inputenc}

\usepackage{times}
\usepackage{latexsym}
\usepackage{amssymb}
\usepackage[T1]{fontenc}
\usepackage[utf8]{inputenc}
\usepackage{microtype}

\usepackage{amsmath}
\usepackage{algpseudocode, algorithmicx, algorithm}
\usepackage{setspace}
\usepackage{multirow}
\usepackage{threeparttable}
\usepackage{graphicx}
\usepackage{booktabs}
\usepackage{makecell}
\usepackage{arydshln}
\usepackage{float}
\usepackage{subfigure}
\usepackage{hyperref}

\newcommand{\hust}{$^1$}
\newcommand{\xiaomi}{$^2$}
\newcommand{\scut}{$^3$}
\newcommand{\ku}{$^4$}

\usepackage[misc]{ifsym}
\newcommand{\cy}[1]{\textcolor{black}{#1}}

\title{Pay More Attention to Relation Exploration for \\ Knowledge Base Question Answering}

\author{Yong Cao\hust, Xianzhi Li\hust\footnotemark[2], Huiwen Liu\xiaomi, Wen Dai\xiaomi, Shuai Chen\xiaomi, \\ {\bf Bin Wang\xiaomi, Min Chen\scut, and Daniel Hershcovich\ku} \\
{\hust} Huazhong University of Science and Technology 
{\xiaomi}Xiaomi AI Lab, China. \\
{\scut}School of Computer Science and Engineering, South China University of Technology \\
{\ku}Department of Computer Science, University of Copenhagen \\
\normalsize{\texttt{\{yongcao\_epic,xzli\}@hust.edu.cn, minchen@ieee.org, dh@di.ku.dk}} \\ \normalsize{\texttt{\{liuhuiwen, daiwen, chenshuai3, wangbin11\}@xiaomi.com}}}

\begin{document}
\maketitle
\begin{abstract}
Knowledge base question answering (KBQA) is a challenging task that aims to retrieve correct answers from large-scale knowledge bases. Existing attempts primarily focus on entity representation and final answer reasoning, which results in limited supervision for this task. Moreover, the relations, which empirically determine the reasoning path selection, are not fully considered in recent advancements. In this study, we propose a novel framework, RE-KBQA, that utilizes relations in the knowledge base to enhance entity representation and introduce additional supervision. We explore guidance from relations in three aspects, including (1) distinguishing similar entities by employing a variational graph auto-encoder to learn relation importance; (2) exploring extra supervision by predicting relation distributions as soft labels with a multi-task scheme; (3) designing a relation-guided re-ranking algorithm for post-processing. Experimental results on two benchmark datasets demonstrate the effectiveness and superiority of our framework, improving the F1 score by 5.8\% from 40.5 to 46.3 on CWQ and 5.7\% from 62.8 to 68.5 on WebQSP, better or on par with state-of-the-art methods.
\vspace{-2mm}
\end{abstract}

\renewcommand{\thefootnote}{\fnsymbol{footnote}} 
\footnotetext[2]{Corresponding author.}
\renewcommand{\thefootnote}{\arabic{footnote}} 

\section{Introduction}
\label{sec:Introduction}

\begin{figure}[t]
	\centering
	\includegraphics[width=0.99\columnwidth]{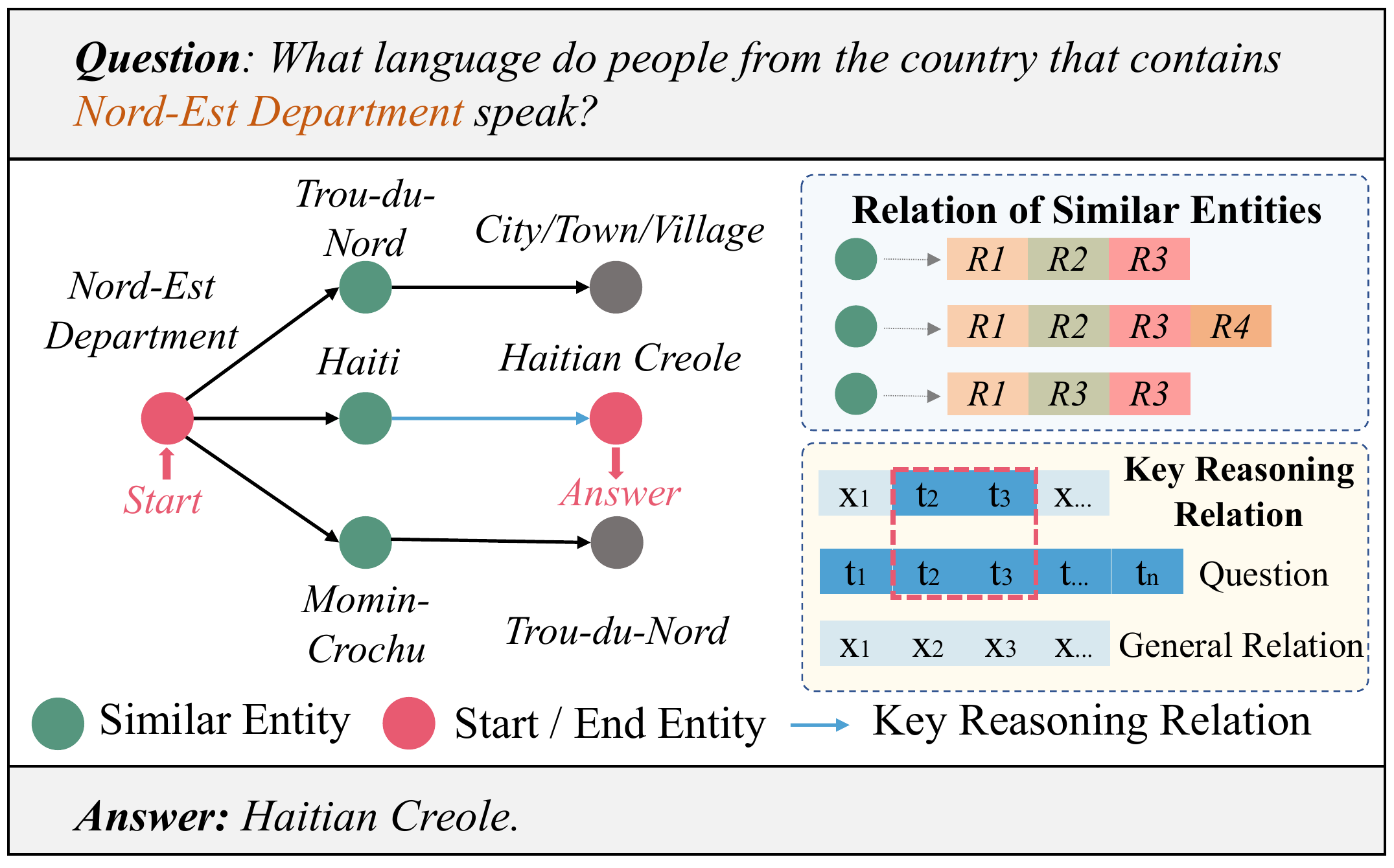}
	\caption{An example of KBQA process. The reasoning begins with the red node and passes through similar entities, which are defined as entities that have similar relations as shown in the upper right box. Besides, key reasoning relations whose tokens ($x_i$) hold overlap between given questions ($t_i$) are important for reasoning.}
	\label{fig:figure1}
\vspace{-5mm}
\end{figure}

Given a question expressed in natural language, knowledge base question answering (KBQA) aims to find the correct answers from a large-scale knowledge base (KB), such as Freebase \cite{bollacker2008freebase}, Wikipedia \cite{vrandevcic2014wikidata}, DBpeidia \cite{auer2007dbpedia}, etc.
For example, the question ``\emph{Who is Emma Stone's father?}'' can be answered by the fact of ``(\emph{Jeff Stone, person.parents, Emma Stone})''.
The deployment of KBQA can significantly enhance a system's knowledge, improving performance for applications such as dialogue systems and search engines.

Early attempts on KBQA \cite{min2013distant, zhang2018variational, xu2019enhancing} mostly focus on transferring given questions into structured logic forms, which are strictly constrained by the consistent structure of parsed query and KB.
To overcome the limitation of the incompleteness of KB, many approaches \citep{xiong2019improving, deng2019multi, lan2021survey} have been developed that aim to map questions and their related KB entities and relations into embeddings, and define the reasoning process as a similarity retrieval problem, which is called IR-based method. Additionally, some studies \cite{ClusterEA, DualMatch, LargeEA22} have attempted to learn relation embeddings and then incorporate surrounding relations to represent entities, which successfully reduces the number of parameters needed for the model.

However, most of these works \citep{han2021two} primarily focus on final answer reasoning and the representation of entities, while few explore the full utilization of relations in KB. Additionally, for answer reasoning, the supervision signal provided is also only from entities, while we believe that the relations also play an important role in determining the reasoning path and the answer choosing.

We propose a new framework, called Relation-Enhanced KBQA (RE-KBQA), to investigate the potential use of relations in KBQA by utilizing an embedding-fused framework. The proposed framework aims to study the role of relations in KBQA in the following three aspects:

\textbf{Relations for entity representation}. We find that similar entities with similar surrounding relations (e.g., the three green circles in the upper right of  Figure \ref{fig:figure1}) play an important role in reasoning. To distinguish them, we introduce QA-VGAE, a question-answering-oriented variational graph auto-encoder, which learns relation weights through global structure features and represents entities by integrating surrounding relations.

\textbf{Relations for extra supervision}. Multi-hop reasoning is often hindered by weak supervision, as models can only receive feedback from final answers \cite{he2021improving}. To overcome this limitation, we propose a multi-task scheme by predicting the relation distribution of the final answers as additional guidance, using the same reasoning architecture and mostly shared parameters. As illustrated in Figure \ref{fig:figure1}, the proposed scheme requires the prediction of both the answer "Haitian Creole" and its surrounding relation distribution.

 \textbf{Relations for post-processing}. We propose a stem-extraction re-ranking (SERR) algorithm to modify the confidence of candidates, motivated by the fact that relations parsed from given questions are empirically associated with strong reasoning paths. As depicted in the bottom of Figure \ref{fig:figure1}, relations that overlap with a given question will be marked as key reasoning relations, and their confidence will be increased empirically. This allows for re-ranking and correction of the final answers.

In general, our contributions can be summarized as follows.
(1) We propose a novel method named Relation Enhanced KBQA (RE-KBQA) by first presenting QA-VGAE for enhanced relation embedding.
(2) We are the first to devise a multi-task scheme to implicitly exploit more supervised signals.
(3) We design a simple yet effective post-processing algorithm to correct the final answers, which can be applied to any IR-based method.
(4) Lastly, we conduct extensive experiments on two challenging benchmarks, WebQSP and CWQ to show the superiority of our RE-KBQA over other competitive methods.
Our code and datasets are publicly available on Github\footnote{\href{https://github.com/yongcaoplus/RE-KBQA}{\nolinkurl{github.com/yongcaoplus/RE-KBQA}}}.
\section{Related Work}
\label{sec:RelatedWorks}

\begin{figure*}[t]
	\centering
	\includegraphics[width=2.1\columnwidth]{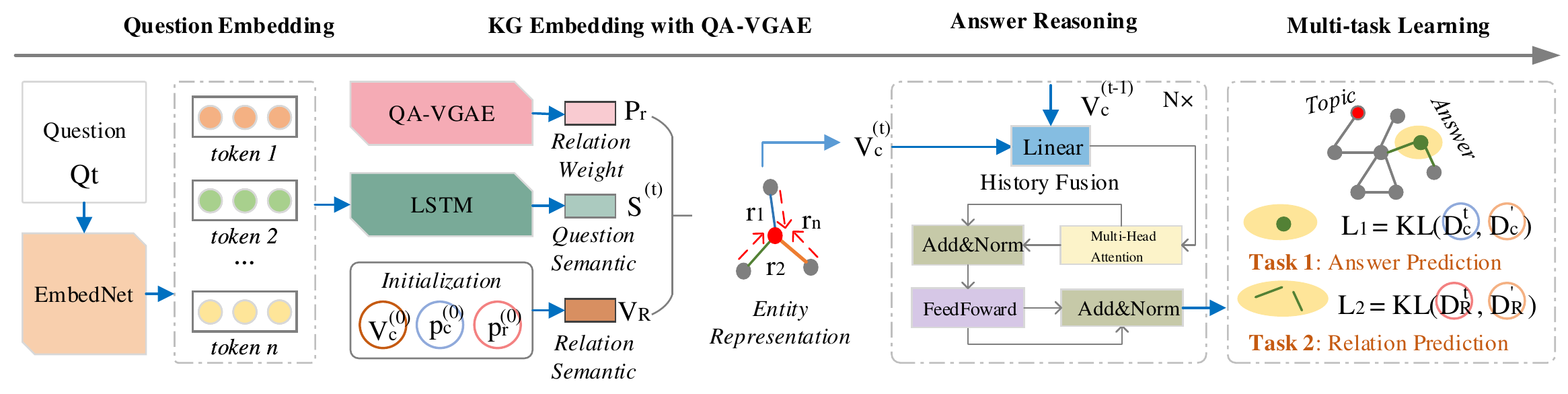}
	\caption{Framework of our proposed approach RE-KBQA. Given a question expressed in natural language, we first employ \emph{question embedding} to encode semantic vectors. Then, we employ \emph{QA-VGAE enhanced representation} module to learn candidate vectors $V_c^{(t)}$, aiming to identify similar entities and key reasoning paths while reasoning. At last, a \emph{multi-task learning} module is proposed to promote training procedure.}
	\label{fig:architecture}
\end{figure*}
\vspace{-3mm}

\paragraph{Knowledge Base Question Answering.}
Most existing research on KBQA can be categorized into two groups: a). Semantic Parsing (SP)-based methods \citep{abdelaziz2021semantic, we2021modeling, cui-etal-2022-compositional}\cy{, which} transfer questions into logical form, e.g., SPARQL queries, by entity extraction, KB grounding, and structured query generation.
b). Information Retrieval (IR)-based method \citep{ding2019leveraging, chen2019uhop, wang2021retrieval, feng2021pretraining,zhang2022improving}\cy{, which} applies retrieve-and-rank mechanism to reason and score all candidates of the subgraph with advancements in representation learning and ranking algorithms.
Apart from the above approaches, recent studies \citep{xiong2019improving, deng2019multi, lan2021survey} also propose several alterations over the reasoning process,
such as extra corpus exploration \citep{xiong2019improving}, better semantic representation \citep{zhu2020knowledge, EASY21},
dynamic representation \citep{han2021two}, and intermediate supervised signals mining \citep{qiu2020stepwise, he2021improving}.
\cy{Aiming to tackle limited corpus, some works are devoted to utilizing external resources, such as using pre-trained language models (Unik-QA) \cite{oguz-etal-2022-unik}, retrieving similar documents (CBR-KBQA) \cite{das2021case}, extra corpus (KQA-Pro) \cite{cao-etal-2022-program}, etc.}

\paragraph{Multi-task Learning for KBQA.} Multitask learning can boost the generalization capability on a primary task by learning additional auxiliary tasks \citep{liu2019self} and sharing the learned parameters among tasks \citep{hwang2021self, xu2021leveraging}. Many recent works have shown impressive results with the help of multi-task learning in many weak supervised tasks such as visual question answering \citep{liang2020learning,rajani2018stacking}, sequence labeling \citep{rei2017semi, yu2021mutually}, text classification \citep{liu2017adversarial, yu2019improving} and semantic parsing \cite{hershcovich-etal-2018-multitask}. In KBQA, auxiliary information is often introduced in the form of artificial ``tasks'' relying on the same data as the main task \citep{hershcovich-etal-2018-multitask, ansari2019neural, gu2021beyond}, rather than independent tasks. This assists the reasoning process and proves to be more effective for the main task. To the best of our knowledge, we are the first to propose a multi-task to assist KBQA by using mostly shared parameters among tasks, for a balance of effectiveness and efficiency.
\section{Problem Formulation}
\label{sec:Problem Formulation}

\paragraph{Knowledge Base (KB).}
A knowledge base usually consists of a huge amount of triples: $\mathcal{G} = \{\langle e, r, e^\prime \rangle | (e, e^\prime) \in \xi, r \in \mathcal{R}\}$,
where $\langle e, r, e^\prime \rangle$ denotes a triple with head entity $e$, relation $r$ and tail entity $e^\prime$.
$\xi$ and $\mathcal{R}$ mean the sets of all entities and relations, respectively.
To apply the triples to downstream task, the entities and relations should be firstly embedded as $d$-dimensional vectors: $V = \{\langle V_e, V_r, V_{e^\prime} \rangle | (V_e, V_{e^\prime}) \in V_{\xi}, V_r \in V_{\mathcal{R}}\} $.

\paragraph{Knowledge Base Question Answering (KBQA).}
Our dataset is formed as question-answer pairs.
Let $Q$ represents the set of given questions and each question $q$ is composed of separated tokens,
where $Q = \{q \in Q | q = x_1, x_2, ..., x_n\}$.
Let $\mathcal{A}$ ($\subseteq \xi$) represents the correct answers of $Q$.
Thus, the dataset is formulated as $\mathcal{D} = \{(Q, A) | (q_1, a_1), (q_2, a_2), ..., (q_m, a_m)\}$.
To reduce the complexity of reasoning process, we extract question-related head entities $e_h$ from $q$ and generate \cy{an associated} subgraph $g_{sub}$ ($\in \mathcal{G}_{sub}$) within multi-hops walking from $e_h$.
Thus, the goal of KBQA is transformed to reason the candidates $c$ ($\subseteq \xi$) of the highest confidence from $g_{sub}$, which can be formalized as:
\begin{equation}\label{eq:fuse}
    c = \mathop{\arg\max}_{\theta, \phi}  {r}_{\phi}(f_{\theta}(q, g_{sub})),
\end{equation}
where $f_{\theta}$($\cdot$) and $r_{\phi}$($\cdot$) denote the representation and reasoning network, respectively.
\section{Our Approach}
\label{sec:The Proposed Approach}
As discussed in Section~\ref{sec:Introduction}, we consider three aspects to further boost the performance of KBQA, including
(i) the enhancement of the representation capability, especially for similar entities;
(ii) a strategy of mining more supervision signals to guide the training; and
(iii) a reasoning path correction algorithm to adjust the ranking results.
Below, we shall elaborate on our network architecture (RE-KBQA) with our solutions to the above issues.

\subsection{Architecture Overview}
\label{sec:kbqa_later_overview}
Inspired by the neighborhood aggregation strategy, we employ Neural State Machine (NSM) \citep{he2021improving} as our backbone model, where entities are denoted by surrounding relations.
We assume that the topic entities and the related subgraph are already achieved by preprocessing; see Section~\ref{subsec:setting} for the details.
Figure~\ref{fig:architecture} shows the main pipeline of our RE-KBQA.
Specifically, given a question $q$, we first employ a question embedding module to encode it into semantic vector.
Here, for a fair comparison with NSM baseline,  we follow \citep{he2021improving} to adopt Glove \citep{pennington2014glove} to encode $q$ into embeddings $\{V_q^j\}_{j=1}^n = \mathrm{Glove}(x_1, x_2, ..., x_n)$, which is then mapped to hidden states by LSTM:
\begin{equation}
	\{h^\prime, \{h_j\}_{j=1}^n\} = \mathrm{LSTM}(V_{q}^1, V_{q}^2, ..., V_{q}^n),
\end{equation}
where we set $h^\prime$ as the last hidden state of LSTM to denote question vector and $\{h_j\}_{j=1}^n$ denotes the vector of tokens.
 After obtaining $h^\prime$ and $\{h_j\}_{j=1}^n$, then we can calculate :
\begin{equation}
  q^{(t)} = \psi (s^{(t-1)}, h^\prime),
\end{equation}
where $\psi$($\cdot$) denotes multi-layer percetron function. Then, the semantic vector $s^{(t)}$ at the $t$-th reasoning step of question $q$ is obtained by:
\begin{equation}
	s^{(t)} = \sum_{j=1}^n p(\psi (q^{(t)}, h_j)) \cdot h_j,
\end{equation}
where $p$($\cdot$) denotes score function, and $s^{(0)}$ ($\in \mathbb{R}^{(|d|)}$) is initialized randomly.

Next, a \emph{QA-VGAE enhanced representation} module is designed to represent KB elements under the guidance of $s^{(t)}$.
Then, unlike previous works that directly predict final answer via a score function, we introduce a \emph{multi-task learning-fused reasoning} module to further predict an auxiliary signal (i.e., relation distribution).
Note that, though we adopt NSM framework to conduct KBQA task, we concentrate on the representation capability enhancement by identifying similar entities, as well as the multi-task learning via supervision signal mining.
At last, to avoid ignoring strong reasoning paths, we further propose a \emph{stem-extraction re-ranking} algorithm to post-process the predictions of our network.
Below, we will present the details of three of our proposed contributed modules.

\subsection{QA-VGAE Enhanced Representation}
\label{sec:qa_vgae}

Similar entities are defined as entities that are connected mostly by the same edges, and only a small portion of edges are different.
For example, as shown in Figure \ref{fig:figure1}, the three nodes marked by dashed circles share almost the same edges, and only the node of ``Haiti'' holds the relation of ``\emph{Person.Spoken\_language}'' that is quite important for answering the question.
Hence, distinguishing similar entities and identifying key reasoning paths are essential for embedding-fused information retrieval-based methods.
Traditional methods like TransE \citep{bordes2013translating} can grasp \cy{local} information from independent triples within a KB, but fail to capture the inter-relations between adjacent triple facts. Consequently, they tend to have difficulties in distinguishing similar entities.

\begin{figure}[t]
  \centering
  \includegraphics[width=0.9\columnwidth]{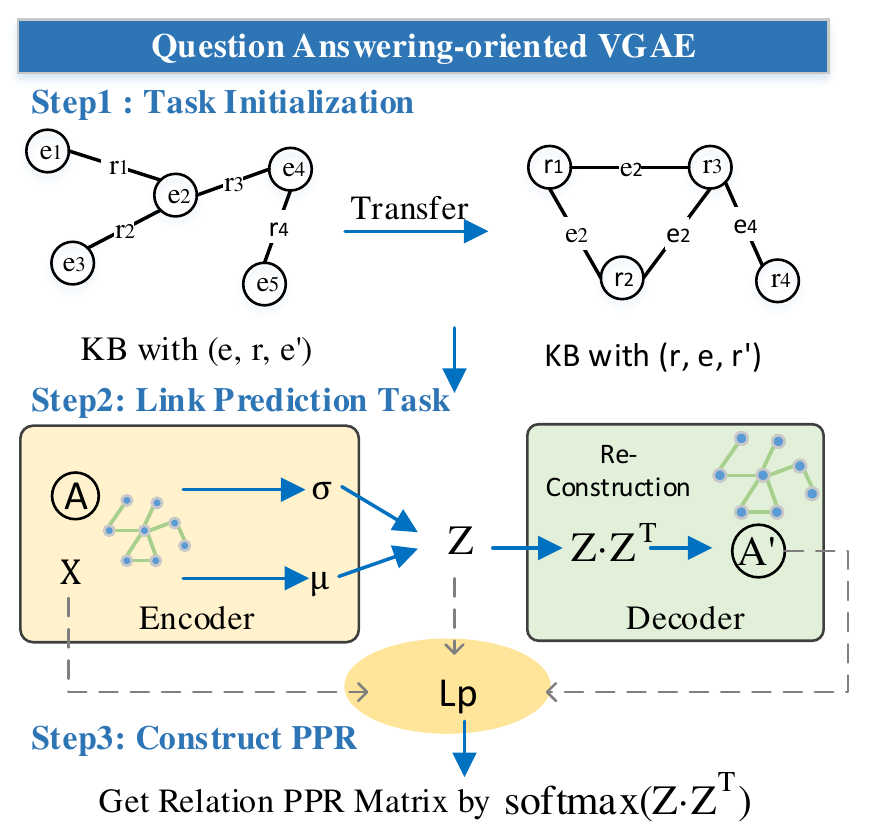}
  \caption{Illustration of training QA-VGAE, including a total of three steps. We adopt two-layers GCN as encoder formalized as $\text{GCN}_{\sigma}$ and $\text{GCN}_{\mu}$.}
  \label{fig:ax_vgae}
  \vspace*{-2mm}
\end{figure}

To alleviate \cy{the above} problem, we introduce  Question Answering-oriented Variational Graph Auto-Encoder (QA-VGAE) module, as is shown in Figure \ref{fig:ax_vgae},  by assigning different weights to reasoning relations, where the weights are learned by VGAE \citep{kipf2016variational}.
Note that, compared with traditional methods like TransE \citep{bordes2013translating}, TransR \citep{lin2015learning}, and ComplEx \citep{trouillon2016complex}, VGAE achieves superior performance in link prediction task.
We thus adopt VGAE in our module to learn weights.
The key insight of this module is to fully learn global structure features by executing graph reconstruction task and constraining the representation as normal distribution, thus promoting the relation representation to be more discriminating.
Finally, by similarity evaluation of the learned representation, we can obtain the prior probability of relation (PPR) matrix, whose elements denote the conditional probability of relations.


In detail, we first transfer the KB from $ \langle e, r, e^\prime \rangle$ (entity-oriented) to $\langle r, e, r^\prime \rangle$ (relation-oriented).
In this way, we can then learn PPR matrix via a link prediction task by unsupervised learning.
Specifically,  given the connection degrees $X$ ($\in \mathbb{R}^{|n_r| \times |n_r|}$) of a relation  and the adjacency $A$ ($\in \mathbb{R}^{|n_r| \times |n_r|}$) between relation nodes, where $n_r$ denotes the number of relations, we adopt two-layers GCN to learn the mean $\sigma$ and variance $\mu$ of the relation importance distribution, and further compound the relation representation $Z$ as :
\begin{equation}
Z = \mathrm{GCN_{\mu}}(X, A) \oplus \mathrm{GCN_{\sigma}}(X, A),
\label{eq:gcn_z}
\end{equation}
where $\oplus$ is compound function. Then, PPR matrix $\mathcal{P}_r$ is obtained by distribution similarity evaluation:
\begin{equation}
\mathcal{P}_r = \mathrm{Softmax} \bigl(Z \cdot Z^\top\bigr),
\label{eq:gae}
\end{equation}
where $\mathcal{P}_r \in \mathbb{R}^{|n_r| \times |n_r|}$. Please refer to Appendix \ref{ax:vgae_details} for loss function $\mathcal{L}_P$ of QA-VGAE.
Next, we denote KB elements as $d$-dim vectors, $V_{\xi} (\in $$\mathbb{R}^{|n_e| \times |d|})$ as entity vectors and $V_{\mathcal{R}} (\in \mathbb{R}^{|n_r| \times |d|})$ as relation vectors, where $n_e$ is the number of enities.
We denote candidate vectors $V_{\mathcal{C}}$ as:
\begin{equation}
	V_{\mathcal{C}} = W_{\mathcal{C}} \cdot \mathcal{P}_r \cdot V_{\mathcal{R}},
	\label{eq4}
\end{equation}
where $W_{\mathcal{C}} \in \mathbb{R}^{|n_c| \times |n_r|}$ denotes \cy{the surrounding relation matrix of entities} and $n_c$ denotes number of candidates.

Then, to integrate semantic vectors $s^{(t)}$ of given question and the history vector, we update $V_c$ as:
\begin{equation}
    \hat{V}_c^{(t)} =  \sigma([V^{(t-1)}_c ; s^{(t)} \odot W_r \odot V_c]),
\end{equation}
where $V_c^{(t)}$  ($\in V_\mathcal{C}$) is candidate vector at time step $t$, $\sigma$($\cdot$) is the linear layer, $[;]$ is the concatenation operation, $\odot$ is element-wise multiplication, and $W_r$ ($\in \mathbb{R}^{|d|}$) is the matrix of learnable parameter.

\subsection{Multi-task Learning-Fused Reasoning}
\label{sec:auxiliary_learning}
The purpose of this module is to conduct answer reasoning from candidate vector $\hat{V}_c^{(t)}$ .
To this end,  we jointly combine the reasoning paths implicitly among candidates by utilizing the Transformer ~\citep{vaswani2017attention}, formalized as:
\begin{equation}
    V^{(t)}_c = \mathrm{Transformer}\left(\left[\hat{V}_{c_1}^{(t)}; \hat{V}_{c_2}^{(t)}; ...; \hat{V}_{c_l}^{(t)}\right]\right),
\end{equation}
where $\{\hat{V}^{(t)}_{c_i}\}_{i=1}^l$ denotes all the candidate vectors.

However, like most existing works \citep{deng2019multi, lange2010deep}, learning from the final answers as the feedback tends to make the model hard to train, due to the limited supervision.
How to introduce extra supervision signals into network model is still an \cy{open question}.
In our method, we introduce a new multi-task to learn the distribution of candidates' surrounding relations, namely surrounding relations reasoning.
The key idea is to leverage relations around final answer as extra supervisions to promote the performance, and also modify reasoning paths implicitly.

Specifically, motivated by weakly-supervised learning methods, we assume the reasoning process starts from topic entity's surrounding relations $S_R^{(0)}$ (initialized along with subgraph generation), and during reasoning, we can easily obtain next surrounding relations' distribution by:
\begin{equation}
    S^{(t)}_R = \sigma \left(\left[(s^{(t)} \cdot V_\mathcal{R}^{\top(t)} ; S_R^{(t-1)}\right]\right),
\end{equation}
where $S^{(t)}_R$ denotes the surrounding relations of candidates at step $t$ and $V_\mathcal{R}^{\top(t)}$ is the transpose of $V_\mathcal{R}$ at step $t$.
Note that, introducing the multi-task will not increase the complexity of our method obviously, since the number of relations is far fewer than that of entities in most cases, and the multi-task shares most parameters with the main task.

In this way, there are two optimization goals of KBQA task, i.e., correct answer retrieving and surrounding relations prediction.
We predict the final answers' possibilities by:
\begin{equation}
    p_c^{(t)} = \mathrm{Softmax} \left(V^{(t)}_c \cdot W_c^{(t)}\right),
\end{equation}
where $p_c^{(t)}$ is the confidence of predicted answers.
Also, the relation distribution confidence $p_r^{(t)}$ is:
\begin{equation}
    p_r^{(t)} = \mathrm{Softmax} \left(S_R^{(t)} \cdot W_r^{(t)}\right),
\end{equation}
where $W_c^{(t)}$ and $W_r^{(t)}$ are learnable parameters.

Then, the answer retriving loss $\mathcal{L}_c$ and the relation prediction loss $\mathcal{L}_r$ can be calculated by:
\begin{equation}
	\begin{aligned}
		\mathcal{L}_c =& \mathrm{KL} ({p}^{(t)}_c, {p}^{(*)}_c) \\
		\mathcal{L}_r =& \mathrm{KL} ({p}^{(t)}_r, {p}^{(*)}_r),
	\end{aligned}
\end{equation}
where ${p_c}^{(*)}$ and ${p_r}^{(*)}$ denote the ground truths, KL is the KL divergence.
Thus, the final total loss is:
\begin{equation}
    \mathcal{L} = \lambda \mathcal{L}_c + (1 - \lambda) \mathcal{L}_r,
\end{equation}
where $\lambda$ denotes a hyper-parameter.

\begin{figure}[t]
	\centering
	\includegraphics[width=0.9\columnwidth]{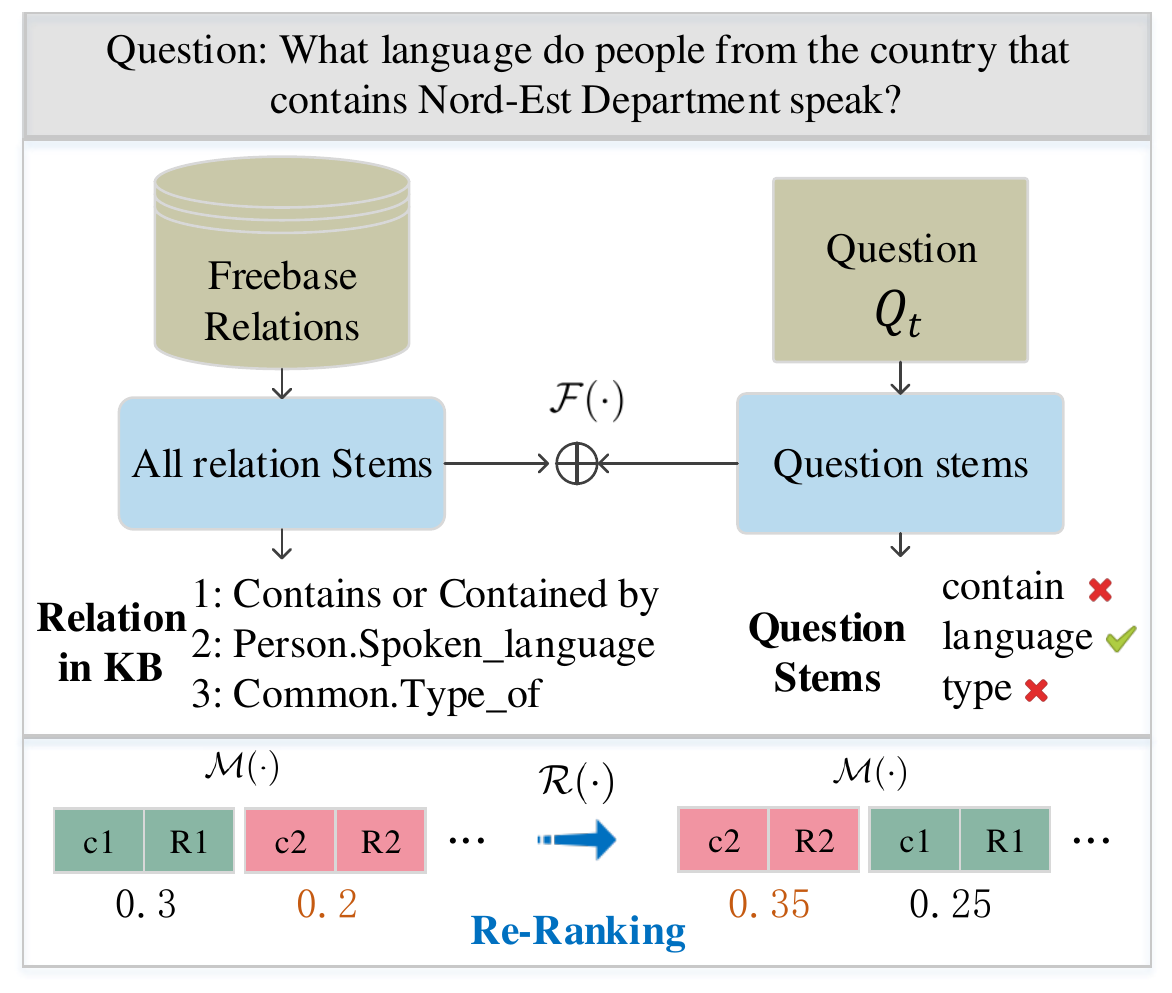}
	\caption{Illustration of SERR algorithm, where stem match mechanism is introduced between KB relations and given questions. If key reasoning relations exist, the rank of candidates will be increased.}
	\label{fig:ax_serr}
	\vspace*{-4mm}
\end{figure}

\subsection{Stem-Extraction Re-Ranking}
\label{sec:serr_algo}
A limitation of embedding-fused KBQA methods is that the reasoning path is uncontrollable as the complete reasoning path is a blackbox in information retrieval-based methods. For example, in the question ``\emph{What is the Milwaukee Brewers mascot?}", the strongly related path ``\emph{education.mascot}"  may be missed due to limited representation capability.
However, this weakness can be easily addressed by semantic parsing-based methods by analyzing the semantic similarity of key elements of questions and relations and constraining the reasoning path. Inspired by this observation, we propose a stem-extraction re-ranking (SERR) algorithm for post-processing. The key idea is to stem-match and re-rank the candidates after obtaining candidates and their confidence from our network.

In detail, we design three operators to execute the re-ranking as shown in Algorithm \ref{alg:SERR}: stemmer $\mathcal{F(\cdot)}$, modifier $\mathcal{M}(\cdot)$, and re-ranker $\mathcal{R(\cdot)}$. These operators are used to extract stems from relations or given questions, modify candidates' confidence, and then re-rank the candidates. As shown in Figure \ref{fig:ax_serr}, given question and candidate predictions, we first use $\mathcal{F(\cdot)}$ to process all the relations of freebase relations and questions. Then, we generate a relation candidates pool by matching the stem pool of the question with the relation stems. This allows us to compare the subgraph of the given question with pseudo-facts produced by given topic entities and candidates, respectively. Finally, according to the comparison, $\mathcal{M}(\cdot)$ and $\mathcal{R(\cdot)}$ are employed to conduct the re-ranking process. 

\begin{algorithm}[ht]
\caption{Stem Extraction Re-Ranking\label{alg:SERR}}
\textbf{Input:} natural language question $Q$, candidates $\mathcal{C}$, confidence $p_{\mathcal{C}}$, relation set $\mathcal{R}$. \\
\textbf{Output:} updated candidates $\mathcal{C}^\prime$ and confidence $p_{\mathcal{C}}^\prime$.
\begin{algorithmic}[1]
\State <* Step 1: Build Relation Trie $\mathcal{P}_s$ *>
\State $\emptyset \rightarrow \mathcal{P}_s$
\ForAll {$r$ in $\mathcal{R}$}
    \State $\mathrm{index}~i, \mathrm{stem}~s = \mathcal{F}(r) $
    \State $\mathcal{P}_s.update(\langle i, s \rangle)$
\EndFor
\ForAll {$\{q, c, p_c\}$ in \{$Q$, $\mathcal{C}$, $p_{\mathcal{C}}$\}}
    \State <* Step 2: Extract Stem of $Q$*>
    \State tokenize $q$ $\rightarrow \mathcal{P}_q$
    \State $\mathcal{F}(\mathcal{P}_q) \rightarrow \mathcal{P}_{stem}^e$
    \State <* Step 3: Re-Ranking $c$ and $p_{c}$*>
    \State $ r_c = \mathrm{match} (\mathcal{P}_{stem}^e, \mathcal{P}_s)$
    \State generate $P = \langle e, \mathcal{P}_s(r_c), e^\prime \rangle$
    \State generate $P^\prime = \langle e, \mathcal{P}_s(r_c)\rangle \cup  \langle e^\prime, \mathcal{P}_s(r_c)\rangle$
    \ForAll {$p$ in $P \cup P^\prime$}
        \If {$p$ in $g_{sub}$ and $p$ in $P$}
            \State $\mathcal{M}(p_{c}, h_1)$
        \EndIf
        \If {$p$ in $g_{sub}$ and $p$ in $P^\prime$}
            \State $\mathcal{M}(p_{c}, h_2)$
        \EndIf
    \EndFor
    \State $c^\prime = \mathcal{R}(c)$ and $p_{c}^\prime = \mathcal{R}(p_{c})$
\EndFor
\end{algorithmic}
\end{algorithm}

It is worth noting that, in our work, we directly use stem extraction method rather than similarity calculation to re-rank.
The insight behind this choice is that, it is unnecessary to consider semantic features again, since we have already injected the question semantic information into our encoded semantic vector $s^{(t)}$,
which means that the model is already equipped with semantic clustering capability.
And obviously, stem extraction costs fewer computation resources, as proved in Appendix  \ref{ax:serr_details}.
Also, our SERR can be migrated to other models as a plug-in and independent module.


\section{Experiments and Results}
\label{sec:Experiments and Results}

\subsection{Datasets}
We conduct experiments on two popular benchmark datasets, including WebQuestionSP \citep{yih2015semantic} and ComplexWebQuestions \citep{talmor2018web}.
Specifically, WebQuestionSP (abbr. WebQSP) is composed of simple questions that can be answered within two hops reasoning, which is constructed based on Freebase \citep{bollacker2008freebase}.
In contrast, ComplexWebQuestions (abbr. CWQ) is larger and more complicated, where the answers require multi-hop reasoning over several KB facts.
The detailed statistics of the two datasets are summarized in Table \ref{tb:dataset_info}.

\begin{table}[]
	\begin{spacing}{1.1}
	\resizebox{0.48\textwidth}{!}{
		\begin{tabular}{l|ccccc}
			\Xhline{1pt}
			Dataset  &  Train  & Valid & Test & \textcolor{black}{Entities} & \textcolor{black}{Relations} \\
			\hline \hline
			WebQSP    & 2,848  & 250   & 1,639  & 259,862   & 6,105 \\
			CWQ       & 27,639 & 3,519 & 3,531  & 598,564   & 6,649 \\
			\Xhline{1pt}
	\end{tabular}}
	\caption{\label{tb:dataset_info} Statistics of WebQSP and CWQ datasets. Note that, \textcolor{black}{\emph{Entities} and \emph{Relations} denote all the entities and relations covered in the subgraph respectively.}}
	\vspace*{-4mm}
	\end{spacing}
\end{table}

\subsection{Experimental Setting}
\label{subsec:setting}
\paragraph{Basic setting.}
To make a fair comparison with other methods, we follow existing works \citep{sun2019pullnet, sun2018open, he2021improving} to process datasets, including candidates generation by PageRank-Nibble algorithm and subgraph construction within three-hops by retrieving from topic entities.
We set the learning rate as $8e-4$ and decay it linearly throughout iterations on both datasets.
We set the number of training epoch on WebQSP and CWQ as 200 and 100, respectively.
For better reproducibility, we give all the parameter settings in Appendix \ref{ax:hyper_param}.

\paragraph{Baselines.}
We compare our method with multiple representative methods, including semantic parsing (SP)-based methods and information retrieval (IR)-based methods.
SPARQA \citep{sun2020sparqa} and QGG \citep{lan-jiang-2020-query} belong to the former category, which focuses on generating optimal query structures.
Besides, KV-Mem \citep{miller2016key}, EmbedKGQA \citep{saxena2020improving}, GraftNet \citep{sun2018open}, PullNet \citep{sun2019pullnet}, ReTraCk \citep{chen-etal-2021-retrack} and BiNSM \citep{he2021improving} are all IR-based methods, which are also the focus of our comparison.

\begin{table}[t]
	\begin{spacing}{1.15}
	\resizebox{0.48\textwidth}{!}{
		\begin{tabular}{lcccc}
			\Xhline{1pt}
			\multirow{2}{*}{Models}&\multicolumn{2}{c}{WebQSP}&\multicolumn{2}{c}{CWQ}\\ \cline{2-5}
			& Hits@1 & F1 & Hits@1 & F1 \\
			\hline  \hline 
			\multicolumn{5}{c}{\emph{SP-Based Method}} \\  \hline
			SPARQA* \citep{sun2020sparqa} & - & - & 31.6 & - \\
			QGG* \citep{lan-jiang-2020-query} & \textcolor{black}{-} & \textcolor{black}{74.0}  & \textcolor{black}{44.1}  & \textcolor{black}{40.4} \\ 
               GNN-KBQA* \citep{hou2022novel} & 68.5 & 68.9 & - & -  \\ \hline
			\multicolumn{5}{c}{\emph{IR-Based Method}} \\ \hline
			KV-Mem{$^\dagger$} \citep{miller2016key}  & 46.6 & 34.5 & 18.4 & 15.7 \\
			EmbKGQA{$^\dagger$} (Saxena et al. 2020) & \textcolor{black}{66.6} & -    & 32.0 &  -   \\
			GraftNet{$^\dagger$} \citep{sun2018open}  & 66.4 & 60.4 & 36.8 & 32.7 \\
			PullNet* (Sun et al. 2019)   & 68.1 & -    & 45.9 & -    \\
			\cy{ReTraCk}{$^*$} \cite{chen-etal-2021-retrack}  & 71.6 & 71.0 & - & - \\
                NSM{$^\dagger$} \citep{he2021improving} & 68.5 & 62.8 & 46.3 &  42.4 \\
			BiNSM* \citep{he2021improving}    & 74.3 & 67.4 & 48.8 & 44.0 \\  
            SR-KBQA* \cite{zhang-etal-2022-subgraph} & 69.5 & 64.1 &  50.2 & \textbf{47.1} \\
            \textcolor{black}{RNG-KBQA*} \cite{ye-etal-2022-rng} & \textcolor{black}{-} & \textcolor{black}{\textbf{75.6}} & \textcolor{black}{-} & \textcolor{black}{-} \\
			\hline 
			\multicolumn{5}{c}{\emph{Ours}} \\ \hline
			RE-KBQA$_{_b}$ & 68.7 & 62.8 & 46.8 & 40.5 \\
			RE-KBQA  & \textbf{74.6} & 68.5 & \textbf{50.3} & 46.3 \\ 
			\Xhline{1pt}
	\end{tabular}}
	\caption{\label{tb:main_results} Performance comparison over state-of-the-art IR-based approaches on WebQSP and CWQ datasets, where bold fonts denote the best scores, * denotes scores from original paper and $\dagger$ are from \citet{zhang-etal-2022-subgraph}.}
	\end{spacing}
 \vspace*{-4mm}
\end{table}

\paragraph{Evaluation metrics.}
To fully evaluate KBQA performance, we should compare both the retrieved and ranked candidates with correct answers.
To this end, we employ the commonly-used F1 score and Hit@1.
F1 score measures whether the retrieved candidates are correct, while Hit@1 evaluates whether the ranked candidate of the highest confidence is in answer sets.

\subsection{Comparison with Others}
\label{subsec:comparing}
We first compare our RE-KBQA against the aforementioned baselines on two datasets and the results are reported in Table~\ref{tb:main_results}.
Note that, RE-KBQA$_{b}$ indicates our backbone network without three modules, i.e., QA-VGAE, multi-task learning and SERR.
Clearly, even using our backbone network, it already outperforms most baselines on two datasets, which is benefited from the semantic guidance of given questions and the reasoning mechanisms.
Further, as shown in the bottom row, our full pipeline achieves the highest values on both datasets over both evaluation metrics.

Particularly,  compared with the results produced by RE-KBQA$_{b}$, our full method improves more on CWQ dataset, which has increased by 3.5 and 5.8 in terms of Hit@1 and F1,
showing that our contributions can indeed boost the multi-hop reasoning process.
Besides, RE-KBQA also obtains good results on simple questions (i.e., WebQSP dataset), especially a 5.7 increase in F1 score, which reveals that the model can recall more effective candidates.

As shown in Table~\ref{tb:main_results}, we can observe that the SP-based methods (i.e., SPARQA and QGG) show a good performance in WebQSP, but perform worse in complicated questions, which reveals that SP-based methods are still weak in multi-hop reasoning.
Similarly, traditional embedding methods, i.e., KV-Mem, EmbedKGQA, and GraftNet, also perform better in simple questions than in complex ones.
Though PullNet and BiNSM show good multi-hop reasoning capacity, the extra corpora analysis and bi-directional reasoning mechanism inevitably increase the complexity of these networks.

Apart from above methods, some attempts are conducted on utilizing additional resources for task enhancement recently. 
As shown in Table~\ref{tb:main_results} \emph{reference}, CBR-KBQA relies on expensive large-scale extra human annotations and Roberta pre-trained model (PLM), Unik-QA tries to retrieve one-hundred extra context passages for relations in KB and T5-base (PLM), and KQA-Pro uses a large-scale dataset for pre-training with the help of explicit reasoning path annotation.
While promising performance has been achieved through these methods, expensive human annotation costs and model efficiency also need to be concerned.

\linespread{1.8}
\begin{table}[t]
	\begin{spacing}{1.1}
		\resizebox{0.48\textwidth}{!}{
			\begin{tabular}{l||cc|cc}
				\Xhline{1pt}
				\multirow{2}{*}{Different cases}&\multicolumn{2}{c|}{WebQSP}&\multicolumn{2}{c}{CWQ}\\ \cline{2-5}
				& Hits@1 & F1 & Hits@1 & F1 \\
				\hline  \hline
				RE-KBQA$_b$  & 68.7
				& 62.8
				& 46.8
				& 40.5 \\
				\hline
				\multirow{2}{*}{with QA-VGAE}   & 73.4
				& 67.7
				& 48.2
				& 45.0 \\
				& \cellcolor[HTML]{B9F4B9} 4.7  $ ^\uparrow $ &  \cellcolor[HTML]{B9F4B9} 4.9  $ ^\uparrow $
				& \cellcolor[HTML]{D8F2D8} 1.4  $ ^\uparrow $ &  \cellcolor[HTML]{B9F4B9} 4.5  $ ^\uparrow $  \\ \hline
				\multirow{2}{*}{with AxLr}  & 72.4
				& 68.4
				& 47.7
				& 42.5 \\
				& \cellcolor[HTML]{C9F3C9} 3.7  $ ^\uparrow $ &  \cellcolor[HTML]{A9F5A9} 5.6  $ ^\uparrow $
				& \cellcolor[HTML]{E7F0E7} 0.9  $ ^\uparrow $ &  \cellcolor[HTML]{D8F2D8} 2.0  $ ^\uparrow $  \\ \hline
				\multirow{2}{*}{with SERR}  & 72.0
				& 65.5
				& 47.3
				& 41.5 \\
				& \cellcolor[HTML]{C9F3C9} 3.3  $ ^\uparrow $ &  \cellcolor[HTML]{C9F3C9} 2.7  $ ^\uparrow $
				& \cellcolor[HTML]{E7F0E7} 0.5  $ ^\uparrow $ &  \cellcolor[HTML]{C9F3C9} 1.0  $ ^\uparrow $  \\ \hline
				\hline
				\multirow{2}{*}{RE-KBQA}  & \textbf{74.6}
				& \textbf{68.5}
				& \textbf{50.3}
				& \textbf{46.3} \\
				& \cellcolor[HTML]{A9F5A9} 5.9  $ ^\uparrow $ &  \cellcolor[HTML]{A9F5A9} 5.7  $ ^\uparrow $
				& \cellcolor[HTML]{C9F3C9} 3.5  $ ^\uparrow $ &  \cellcolor[HTML]{A9F5A9} 5.8  $ ^\uparrow $  \\
				\Xhline{1pt}
		\end{tabular}}
		\caption{\label{tb:main_ablation} Comparing our full pipeline (bottom row) with various cases in the ablation study.
			The cells with different background colors reveal the improvement over our backbone network RE-KBQA$_b$.}
	\end{spacing}
	\vspace*{-3mm}
\end{table}

\subsection{Network Component Analysis}
To evaluate the effectiveness of each major component in our method, we conducted a comprehensive ablation study.
In detail, similar to Section~\ref{subsec:comparing}, we remove all three components and denote the backbone network as RE-KBQA$_b$.
Then, we add QA-VGAE (Section~\ref{sec:qa_vgae}), multi-task learning (Section~\ref{sec:auxiliary_learning}), and SERR (Section~\ref{sec:serr_algo}) back on RE-KBQA$_b$, respectively.
In this way, we constructed totally four network models and re-trained each model separately using the same settings of our RE-KBQA model.
Table \ref{tb:main_ablation} shows the results. By comparing different cases with the bottom-most row (our full pipeline), we can see that each component contributes to improving the performance on both datasets.
More ablation experiments can be found in Appendix.
Below, we shall discuss the effect of each module separately.

\begin{figure}[t]
	\centering
	\includegraphics[width=1.0\columnwidth]{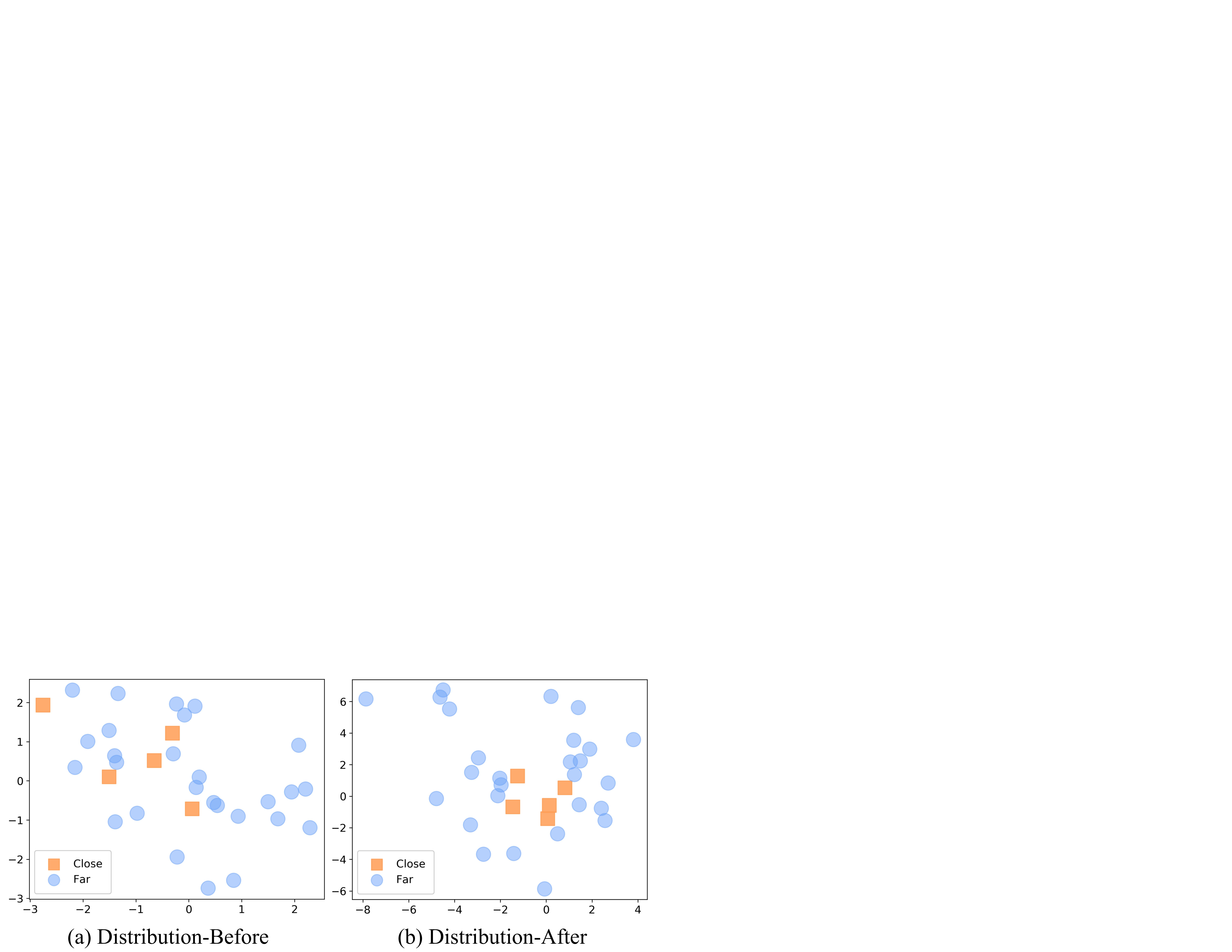}
	\caption{Relation vector visualization in the case of \emph{``What is the capital of Austria?''} via T-SNE. Orange nodes indicate relations close to ``\emph{location}'' and blue nodes indicate far relations.}
	\label{fig:relation_visual}
\end{figure}

\paragraph{Effect of QA-VGAE.} \
From the results of Table \ref{tb:main_ablation}, we can observe that the improvements of using QA-VGAE are more remarkable than using the other two modules, demonstrating that the QA-VGAE is more helpful to boost the reasoning process for both simple and complex questions.
Besides quantitative comparison, we also tried to reveal its effect in a visual manner.
Here, we adopt T-SNE to visualize the relation vectors.
Figure \ref{fig:relation_visual} shows a typical embedding distribution before and after QA-VGAE training.
For a clear visualization, we randomly select some relations related to a case ``\emph{What is the capital of Austria?}''.
The orange nodes represent relations close to ``\emph{location}'', such as ``\emph{location.country.capital}'', ``\emph{location.country.first\_level\_divisions}'', etc.,
and the blue nodes denote the relations that are not covered by the question subgraph, which we call far relations.
Obviously, after using QA-VGAE, the related relations (orange nodes in (b)) tend to get closer and the other nodes get farther.

\begin{figure}[t]
	\centering
	\includegraphics[width=1.0\columnwidth]{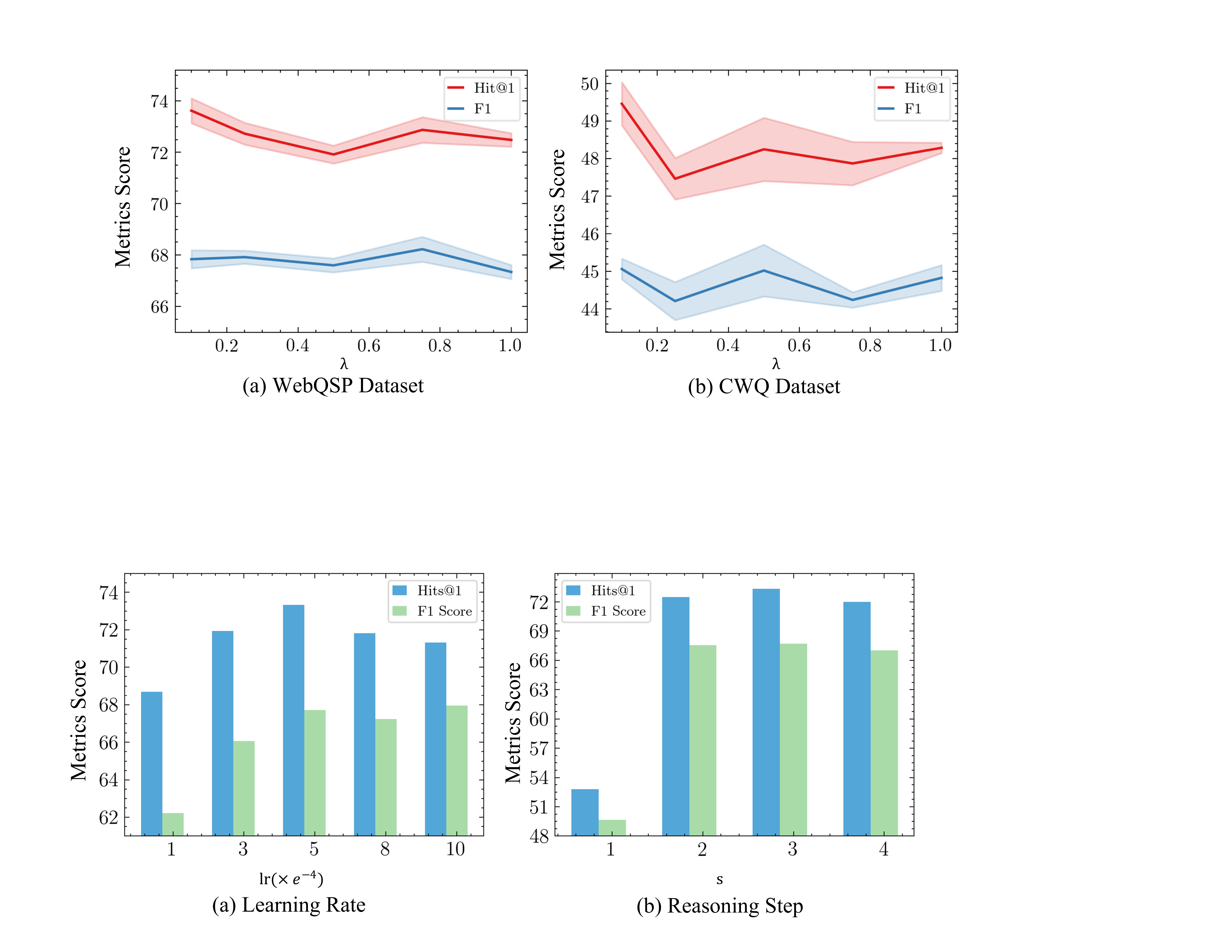}
	\caption{Analysis of using different loss fusion weights among two benchmark test sets in multi-task learning.}
	\label{fig:ablation}
\end{figure}

\paragraph{Effect of multi-task learning.} \
As shown in Table \ref{tb:main_ablation}, the multi-learning module shows better performance in simple questions (see WebQSP dataset), since the relation distribution is denser than candidates distribution, thus causing the prediction to be more complicated along with the increase of reasoning steps.
To fully explore the effect of this module, we study different loss fusion weights and the results are shown in Figure \ref{fig:ablation}, where a larger $\lambda$ (range from 0.1 to 1.0, and we discard the setting of 0.0 for its bad performance) denotes a more weighted loss of main task.
Clearly, only designing the primary task or auxiliary task is not optimal for KBQA, and the best setting of $\lambda$ is 0.1 and 0.5 for the two datasets. 
\cy{An interesting observation is that the best Hit@1 is obtained with lower lambda while the best F1 score is obtained with higher lambda in each dataset. 
We claim that it is caused by the different goals of Hit@1 and F1 metrics, that is, Hit@1 shows whether the top one candidate is found while F1 score evaluates whether most candidates are found.}

\begin{figure}[ht]
        \centering
        \subfigure[Case analysis for SERR with the question of ``\emph{What Chamorro Time Zone countries have territories in Oceania?}''. To be clear, we just show top three representative paths here.]{
            \begin{minipage}[b]{0.96\columnwidth}
            \includegraphics[width=0.985\columnwidth]{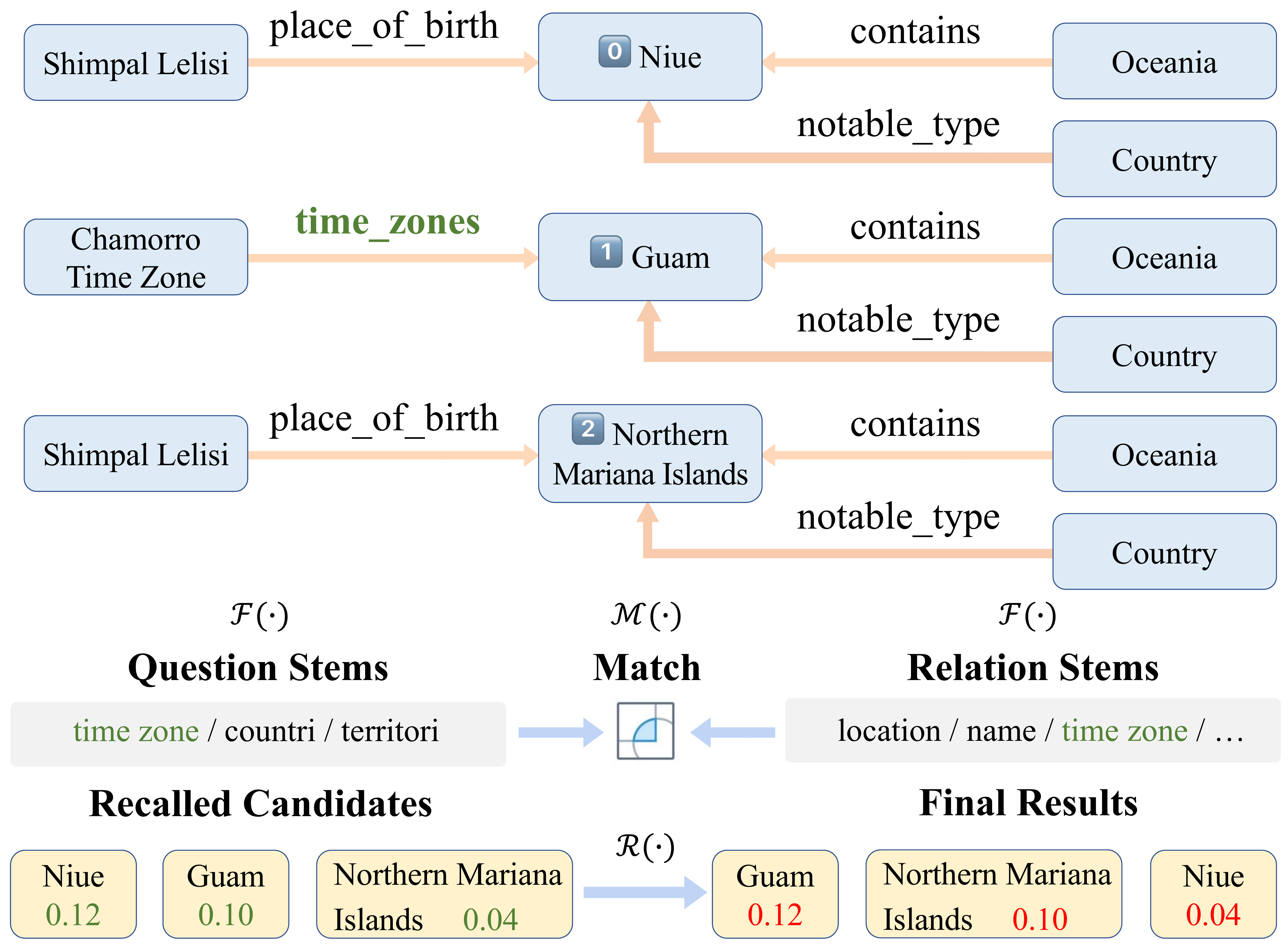}
            \end{minipage}
            \label{fig:serr_case1}
            }
        \subfigure[Case analysis for RE-KBQA with proposed modules.]{
            \begin{minipage}[b]{0.98\columnwidth}
            \includegraphics[width=1.0\columnwidth]{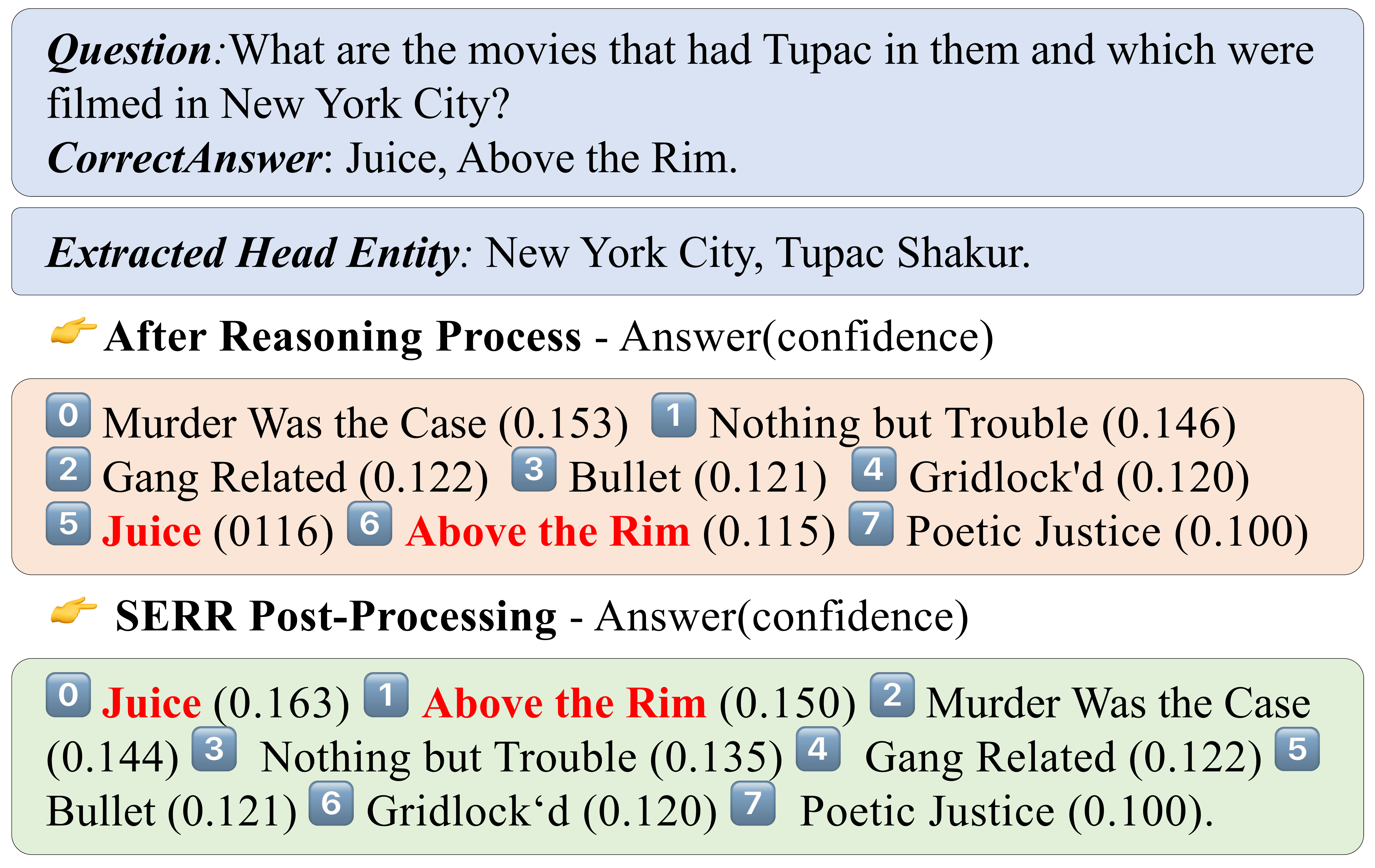}
            \end{minipage}
            \label{fig:serr_case2}
            }
        \caption{Case analysis of multi-hop reasoning process.}
        \label{fig:serr_case}
\end{figure}

\paragraph{Effect of SERR.} \
This module is lightweight (see Appendix \ref{ax:serr_details} for inference time) yet effective, especially for simple questions; see Table \ref{tb:main_ablation}.
Intuitively, the stem extraction for key paths is quite effective for questions that rely on direct-connected facts.
In contrast, stem extraction for complex questions relies more on the startpoint and endpoint.
Figure \ref{fig:serr_case1} further shows an example result of SERR module, which proves that it can effectively identify close connected facts of a given question and re-rank the candidates.

\subsection{Case Study}
At last, we show a case result produced by our RE-KBQA; see Figure \ref{fig:serr_case2}. Given the question ``What are the movies that had Tupac in them and which were filmed in New York City?'',
our method first embedded the question into vectors and retrieve related subgraphs. Then, by utilizing the promotion of our proposed QA-VGAE and multi-task learning, we can use the trained model and obtain the candidates of ``\emph{Murder Was the Case}'', ``\emph{Nothing but Trouble}'', etc,
and thanks to the SERR algorithm, our reasoning process can have a chance to re-rank the candidates, thus boosting its performance.
Finally, we output \emph{Juice} and \emph{Above the Rim} as the correct answers.
For similarity entity identification, \cy{SERR in other methods as a plug-in }and more case results, please refer to Appendix \ref{ax:serr_details} and \ref{ax:more_cases}.

\vspace{-3mm}
\section{Conclusion}
In this paper, we proposed a novel framework, namely RE-KBQA, with three novel modules for knowledge base question answering, which are QA-VGAE to explore the relation promotion for entity representation, multi-task learning to exploit relations for more supervisions, and SERR to post-process relations to re-rank candidates.
Extensive experiments validate the superior performance of our method compared with state-of-the-art IR-based approaches.

\section{Limitations}
\label{sec:Limitations and Future Work}
While good performance has been achieved, there are still limitations in our work.
First, though QA-VGAE extracts enhanced features and are fast to train, it is an independent module from the main framework.
Second, as a post-processing step, the performance of SERR module on simple question is better than that of complex questions.

In the future, we would like to explore the possibility of fusing relation constraints into the representation module directly and inject strong facts identification mechanism as guidance signal of multi-hop reasoning process, aiming to integrate QA-VGAE and SERR into the main framework.

\section*{Acknowledgments}
Thanks to the anonymous reviewers for their helpful feedback. We gratefully acknowledge the insightful suggestions from Zeqi Tan. This work is supported by the China National Natural Science Foundation No. 62202182. Yong Cao is supported by China Scholarship Council (No. 202206160052) and the Zhejiang Lab’s International Talent Fund for Young Professionals.

\bibliography{anthology,custom}
\bibliographystyle{acl_natbib}

\appendix
pdf

\section{Appendix}
\subsection{QA-VGAE Training}
\label{ax:vgae_details}
In this section, we introduce the details of the QA-VGAE training procedure and demonstrate its effectiveness.

\paragraph{Training Goal.}
We adopt encoder-decoder models to conduct relation reconstruction tasks.
Given the prepared adjacent matrix $A$, and feature matrix $X$, we use a two-layer GCN as a distribution learning model to estimate its mean and variance.
The training loss function is formalized as:
\begin{equation}
\mathcal{L}_P =  \mathbb{E}_{q(Z|X,A)}[\log p (A\,|\,Z )] - \mathrm{KL}(q(X, A), p(Z))
\end{equation}
where $Z$ is calculated by Equation \ref{eq:gcn_z},
$\mathrm{KL}$ is the Kullback-Leibler divergence, $q(\cdot)$ and $p(\cdot)$ denotes the encoder and decoder respectively, please refer to \citet{kipf2016variational} for more details.

\paragraph{Settings.}
Specifically, $A$ is defined as the matrix of neighborhood relations between nodes, where we set $A(i, j)$ as 1 if there is a connection between relation $r_i$ and $r_j$, and 0 for no connection.
$X$ is the feature matrix defined as the connectivity, which is accumulated as the number of edges between two nodes, aiming to show the importance of a relation.
We set an empirical thresh of each element in the feature matrix to avoid extremely large values to hurt the model's training, such as the degrees of ``\emph{Common.type\_of}'' is quite huge, defined as:
\begin{equation}
\label{eq6}
X[i, j]=\left\{
\begin{aligned}
\tau & , & c \geq \tau, \\
c & , & c < \tau.
\end{aligned}
\right.
\end{equation}
where $c$ is connectivity, $\tau$ is an emprical hyper-parameter, and we set $\tau$ as $2000$ in our work.


\subsection{SERR Algorithm}
\label{ax:serr_details}

\paragraph{Complexity Analysis.}
Definitely, applying semantic similarity between relations and given questions is a more straightforward method to identify strong relations.
However, the process of such a method is more complicated and time-consuming.
To prove the efficiency of our method, we conduct a comparison experiment to reflect the complexity of the two methods.
As is shown in Table \ref{tb:serr_semantic_compare}, the top two rows denote semantic similarity method, and the last row denotes our method.
Obviously, our method is more lightweight without extra pre-trained models and the dependence on GPU resources.
For comparison, we adopt \emph{Bert-base-uncased} model to conduct the semantic similarity process in this experiment, which can be downloaded in \url{https://huggingface.co/bert-base-uncased}.


\linespread{1.8}
\begin{table}[ht]
\begin{spacing}{1.2}
\resizebox{0.48\textwidth}{!}{
\begin{tabular}{l||ccc|ccccc}
  \Xhline{1pt}
  \multirow{2}{*}{Module}&\multicolumn{3}{c|}{WebQSP}&\multicolumn{3}{c}{CWQ}\\ \cline{2-7}
  & Params & Time & GPU & Params & Time & GPU \\
  \hline  \hline
  Cosine Distance   & 420.10 & 28.8 & $\surd$ & 420.10 & 53.5&    $\surd$    \\
  Euclidean Distance & 420.10 & 30.5 & $\surd$ & 420.10 & 55.5&    $\surd$    \\
  Stem Extraction   & - & 4.9  & $\times$ & - & 18.3 &  $\times$ \\
  \Xhline{1pt}
\end{tabular}}
\caption{\label{tb:serr_semantic_compare} Comparing SERR module with semantic similarity method, i.e., cosine distance and euclidean distance in terms of model parameters and computing resources. \emph{Time} row denotes total handling time (\emph{minutes}). \emph{Params} row denotes model size (\emph{MB})}
\end{spacing}
\end{table}

\paragraph{Performance Analysis.}
Besides, to demonstrate it can be plug-in and infer cases quickly, we further validate its accuracy and inference time, as is shown in Table \ref{tb:serr_performance},
Note that, since SERR relies on traditional stem extraction rather than semantic understanding to identify the key paths, there is no training period for SERR, and it can be applied to any information-retrieval(IR)-based methods.

\cy{Finally, to demonstrate the plug-in attributes of the SERR module, we integrate this module into BiNSM network \cite{he2021improving} and the results are shown in Table \ref{tb:serr_in_binsm}. The results show that SERR can indeed increase the Hit@1/F1 score from 74.3/67.4 to 74.8/68.0 in the WebQSP dataset, and from 48.8/44.0 to 49.5/45.3 in the CWQ dataset.}

\linespread{1.8}
\begin{table}[ht]
		\centering
		\begin{spacing}{1.2}
		\resizebox{0.28\textwidth}{!}{
			\begin{tabular}{l|c|c}
				\Xhline{1pt}
				Factor            & Webqsp & CWQ \\
				\hline \hline
				Accuracy (\%)    &  63.2  & 75.5   \\
				Infer Time (s)   &  0.18  & 0.32       \\
				\Xhline{1pt}
		\end{tabular}}
		\caption{\label{tb:serr_performance} Performance of SERR algorithm in terms of accuracy score and inference time in two benchmark datasets. The accuracy score is calculated among recalled cases where close facts lie in its subgraph.}
		\end{spacing}
\end{table}

\linespread{1.8}
\begin{table}[h]
		\resizebox{0.48\textwidth}{!}{
			\begin{tabular}{l||cc|cc}
				\Xhline{1pt}
				\multirow{2}{*}{Different cases}&\multicolumn{2}{c|}{WebQSP}&\multicolumn{2}{c}{CWQ}\\ \cline{2-5}
				& Hits@1 & F1 & Hits@1 & F1 \\
				\hline  \hline
				BiNSM   & 74.3 & 67.4 & 48.8 & 44.0 \\
				\hline
				\multirow{2}{*}{with SERR }  & 74.8 & 68.0 & 49.5 & 45.3  \\
				&   \cellcolor[HTML]{D8F2D8} 0.5  $ ^\uparrow $ 
				&   \cellcolor[HTML]{D8F2D8} 0.6  $ ^\uparrow $
				&   \cellcolor[HTML]{C9F3C9} 0.7  $ ^\uparrow $
				&   \cellcolor[HTML]{A9F5A9} 1.3  $ ^\uparrow $ \\
				\Xhline{1pt}
		\end{tabular}}
		\caption{\label{tb:serr_in_binsm} Integrating SERR module into BiNSM network to demonstrate it can be a plug-in and independent module for any IR-based methods.
			The cells with different background colors reveal the improvement over our SERR module.}
	\vspace*{-2mm}
\end{table}

\subsection{Hyper-parameter Setting.}
\label{ax:hyper_param}
In order to help reproduce RE-KBQA and its reasoning performance, as shown in Table \ref{tb:para_setting}, we list the hyper-parameters of the best results on two benchmark datasets.
For the WebQSP dataset, the best results are obtained by using the initial learning rate of 0.0008,  training batch size of 40, dropout rate of 0.30, reasoning step of 3, and max epoch size of 100.
For the CWQ dataset, the best results are obtained by using the initial learning rate of 0.0008,  training batch size of 100, dropout rate of 0.30, reasoning step of 3, and max epoch size of 200.
For more experiment details, please refer to our code which will be published upon the publication of this work.

\linespread{1.8}
\begin{table}[htbp]
\centering
\begin{spacing}{1.2}
\resizebox{0.3\textwidth}{!}{
\begin{tabular}{l||l|l}
\Xhline{1pt}
\textbf{Parameter}   & WebQSP & CWQ \\
\hline \hline
\emph{Learning rate} & $8e^{-4}$ & $8e^{-4}$ \\
\emph{Batch size}    & 40        & 100       \\
\emph{Eps}  		  & 0.95      & 0.95      \\
\emph{Dropout}  		  & 0.30      & 0.30      \\
\emph{Num\_step}     & 3         & 3         \\
\emph{Entity\_dim}	  & 50        & 50        \\
\emph{Word\_dim}     & 300       & 300       \\
\emph{Num\_epoch}    & 200       & 100       \\
\emph{Relations}     & 6105      & 6649      \\
\emph{Num\_candidates}     & 2000      & 2000      \\
\Xhline{1pt}
\end{tabular}}
\caption{\label{tb:para_setting} The hyper-parameters of the best results on WebQSP and CWQ dataset for the KBQA task.}
\end{spacing}
\end{table}

\begin{figure}[hbt!]
        \centering
        \subfigure[Learning rate $(\times e-4)$  ablation study of WebQSP dataset.]{
            \begin{minipage}[b]{1.0\columnwidth}
            \includegraphics[width=0.7\columnwidth]{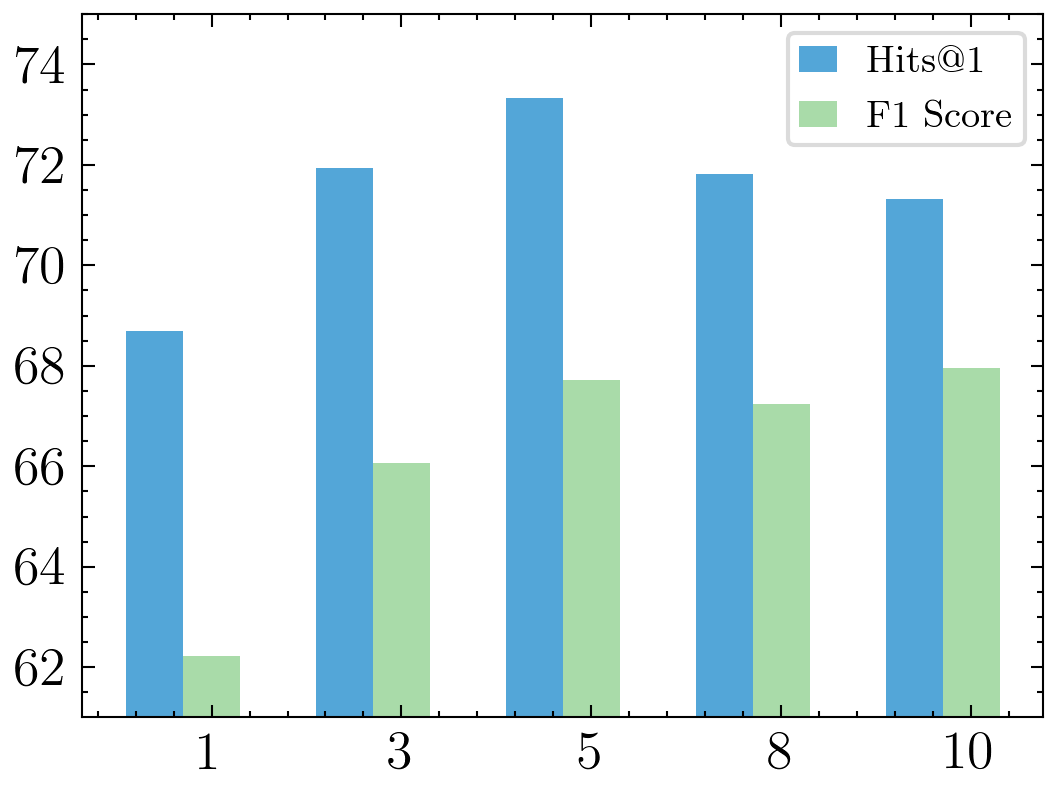}
            \centering
            \end{minipage}
            \label{fig:lr_web}
            }
        \subfigure[Performance comparison over different reasoning steps of WebQSP dataset.]{
            \begin{minipage}[b]{1.0\columnwidth}
            \includegraphics[width=0.7\columnwidth]{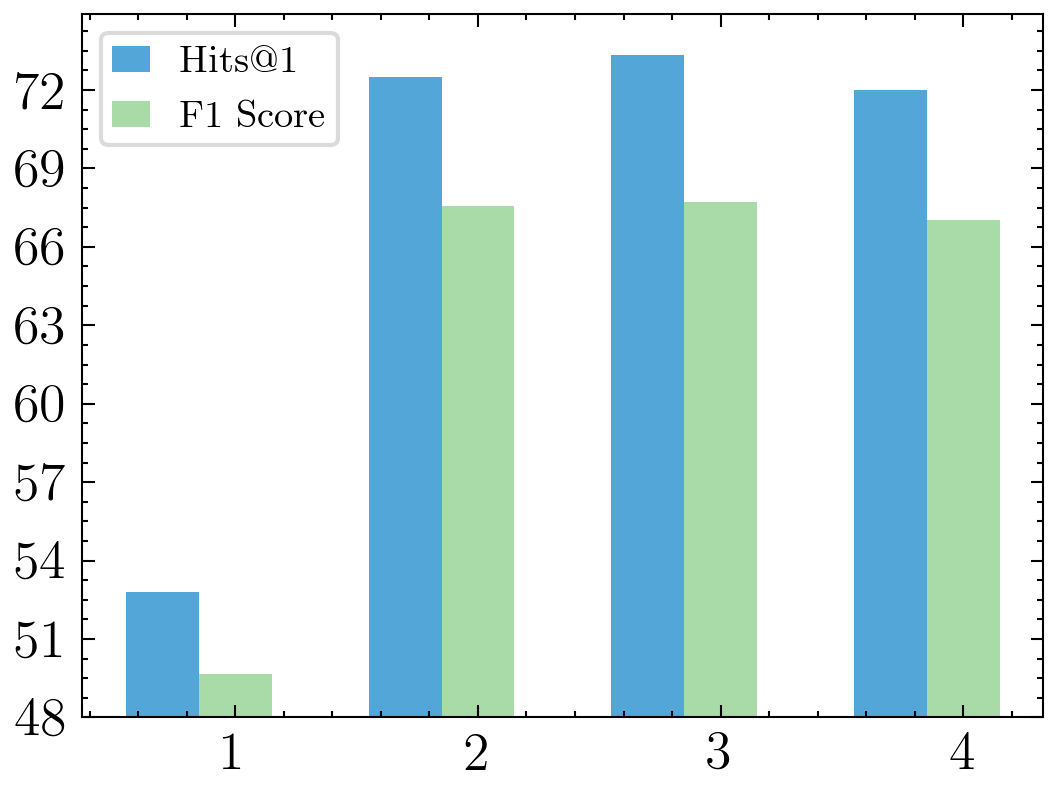}
            \centering
            \end{minipage}
            \label{fig:step_web}
            }
        \caption{More ablation analysis of reasoning process produced by RE-KBQA.}
        \label{fig:serr_case}
\end{figure}

\subsection{More Ablation Study}
\label{ax:more_ablation_study}
\paragraph{Reasoning Network.}
One minor modification of our work is that we adopt Transformer Encoder as a reasoning network, of its self-attention mechanism and superior capability of encoding information.
As is shown in Table \ref{tb:ablation_network}, compared with the backbone model (Linear layer), LSTM can acquire slight performance but with obviously longer training time, and Transformer Encoder can obtain promotion for KBQA task with tolerable extra training time.
Therefore, different reasoning layers also affect the performance, and adopting Transformer Encoder can benefit a lot with three modules.

\paragraph{Training Settings.}
From Figure \ref{fig:lr_web} and  \ref{fig:step_web}, we further study that $5e-8$ and $3$ is the best hyper-parameter setting for the learning rate and reasoning step. It is worth noting that, for embedding-fused methods, the more reasoning steps are not the determinant for network performance. We conduct our experiments on 2* V100 GPUs.

\linespread{1.8}
\begin{table}[ht]
        \begin{spacing}{1.2}
		\resizebox{0.48\textwidth}{!}{
			\begin{tabular}{l||ccc|ccc}
				\Xhline{1pt}
				\multirow{2}{*}{Models}&\multicolumn{3}{c|}{WebQSP}&\multicolumn{3}{c}{CWQ}\\ \cline{2-7}
				& Hits@1 & F1 & Train & Hits@1 & F1 & Train\\
				\hline  \hline
				RE-KBQA$_b$ & 68.7 & 62.8 & 4.3 & 46.8 & 40.5 & 21.1 \\ \hline
				LSTM   & 70.9 & 66.6 & 6.5 & 47.6 & 41.5 & 27.0 \\
				&   2.2  $ ^\uparrow $ &   3.8  $ ^\uparrow $
				& \cellcolor[HTML]{C9F3C9} 2.2  $ ^\uparrow $ &    0.8  $ ^\uparrow $
                &   1.0  $ ^\uparrow $ &  \cellcolor[HTML]{A9F5A9} 5.9  $ ^\uparrow $\\ \hline
				Transformer  & 71.0 & 66.1 & 4.5 & 47.1 & 42.7 & 21.5 \\
				&  2.3  $ ^\uparrow $ &    3.3  $ ^\uparrow $
				& \cellcolor[HTML]{D8F2D8} 0.2  $ ^\uparrow $ &    0.3  $ ^\uparrow $
                &   1.2  $ ^\uparrow $ &   \cellcolor[HTML]{D8F2D8} 0.4  $ ^\uparrow $\\ \hline
				\Xhline{1pt}
		\end{tabular}}
		\caption{\label{tb:ablation_network} Hit@1, F1 score and training time comparison of backbone model with different reasoning networks. \emph{Train} row denotes training time in hours.
        The cells with different background colors reveal the extra training time over our backbone network RE-KBQA$_b$.}
        \end{spacing}
\end{table}

\begin{figure}[ht!]
	\centering
	\includegraphics[width=1.0\columnwidth]{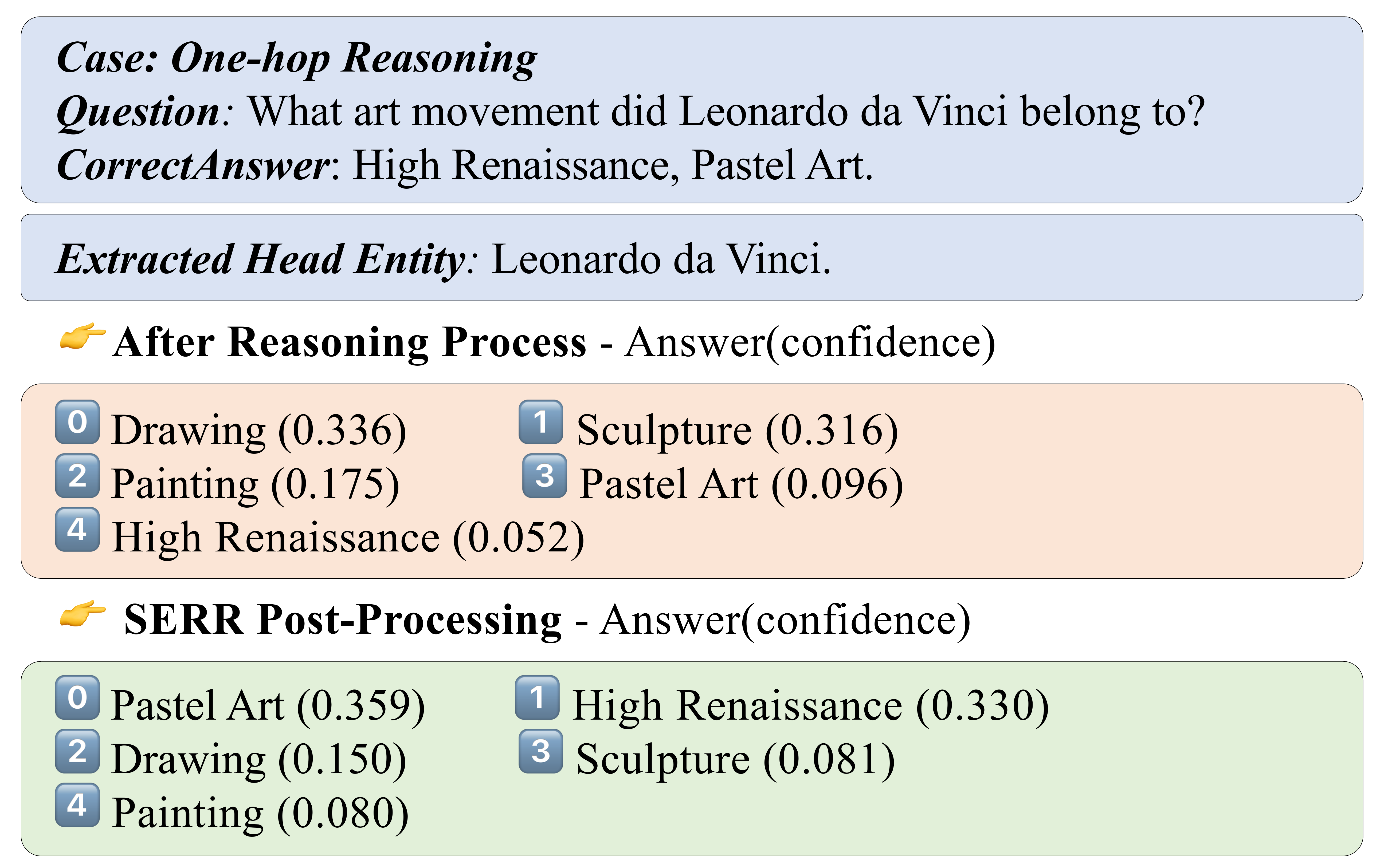}
	\caption{An example of one-hop reasoning process produced by our method. 
	}
\vspace*{-2mm}
	\label{fig:serr_case_single}
\end{figure}

\begin{figure*}[t]
\centering
\subfigure[Similar Entity Identification.]{
\begin{minipage}[t]{0.5\linewidth}
\centering
\includegraphics[width=1\linewidth]{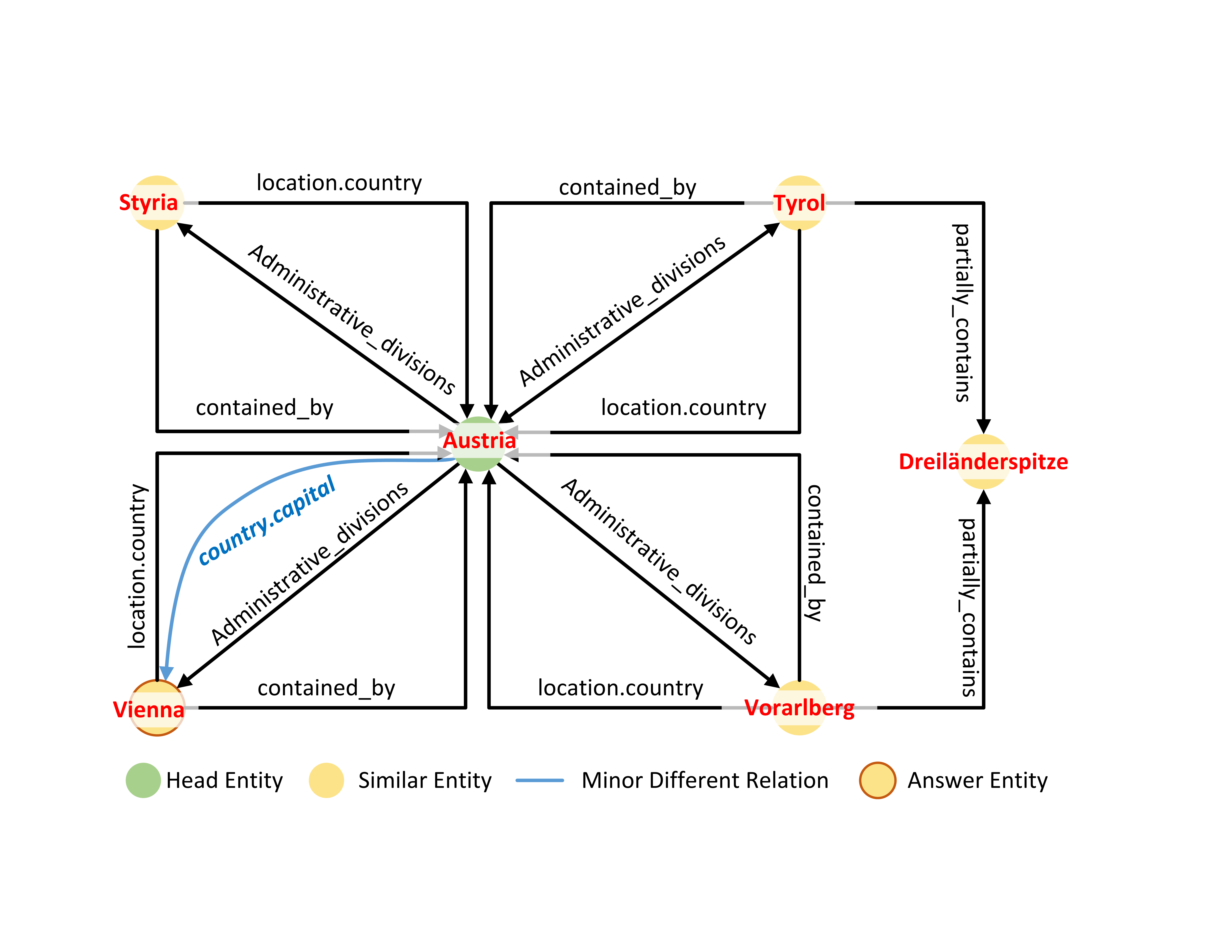}
\label{fig:similari_distinguish}
\end{minipage}%
}%
\subfigure[Reasoning process of RE-KBQA]{
\begin{minipage}[t]{0.5\linewidth}
\centering
\includegraphics[width=1\linewidth]{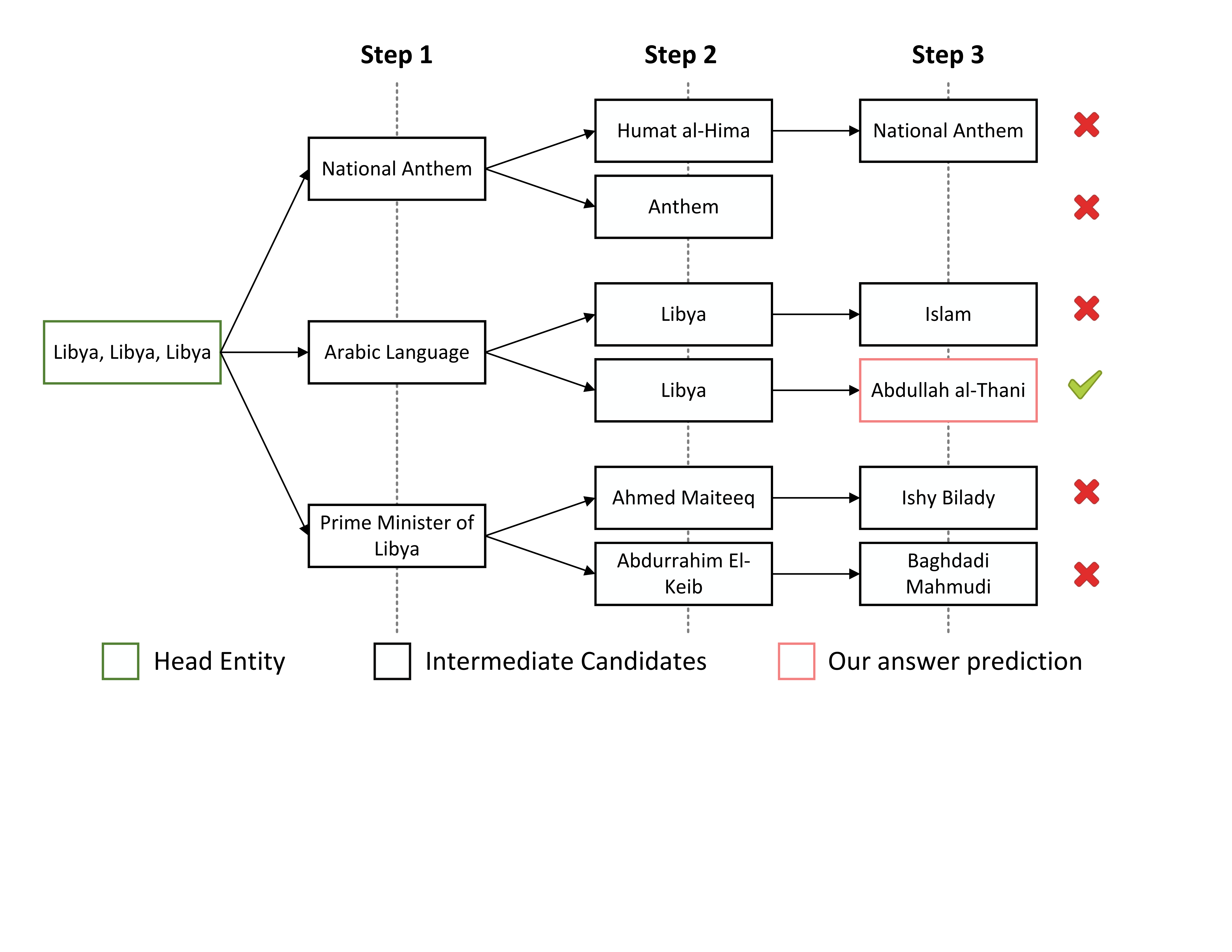}
\label{fig:after_kbqa}
\end{minipage}
}
\caption{Cases Analysis of similar entity and thorough process of RE-KBQA compared with backbone network. Specifically, (a) is to demonstrate that our model can reason correct answers across similar entities that benefited from QA-VGAE in case \emph{``What is the capital of Austria?''}. (b) aims to show the full pipeline of our proposed method in case \emph{``Which man is the leader of the country that uses Libya, Libya, Libya as its national anthem?''}  }.
\end{figure*}

\subsection{More Case Analysis}
\label{ax:more_cases}
In this section, we deliver more case analysis on simple questions, similarity entity identification, and the intuitive reasoning process of our method.
\paragraph{Simple questions.}
As shown in Figure~\ref{fig:serr_case_single}, we show a case of one-hop reasoning on the WebQSP dataset, which proved that RE-KBQA performs well in simple question answering, as the main network can recall correct candidates and the SERR module can effectively re-rank the candidates.

\paragraph{Similarity entity identification.}
To demonstrate our method can indeed distinguish similar entities, we choose a case that needs to reason across similar entities as is shown in Figure\ref{fig:similari_distinguish}. While most of the surrounding edges are the same among candidates of the first step, our method can still select the correct node as the final answer.

\paragraph{RE-KBQA reasoning process.}
Figure~\ref{fig:after_kbqa} shows a three-hop reasoning case of our method, to intuitively demonstrate that our method can effectively conduct a multi-hop reasoning process. Note that, the reasoning process of our method can be illustrated as the status transfer of the relation $V_r^{(t)}$ and candidate vectors $V_c^{(t)}$ from one distribution into another, which is not strictly consistent along the reasoning path, thus in some degree solve the problem of knowledge base incompleteness.

\end{document}